\documentclass[10pt,twocolumn,letterpaper]{article}

\usepackage{cvpr}              % To produce the CAMERA-READY version
\usepackage{tabularx}
 \usepackage{multirow}
 \usepackage[table,xcdraw]{xcolor}
 \usepackage[normalem]{ulem}
 \usepackage{amssymb}
	
\usepackage{graphicx}
\usepackage{amsmath}
\usepackage{amssymb}

\usepackage{booktabs}
\usepackage{enumitem}

\newcolumntype{Y}{>{\RaggedRight\arraybackslash}X} 
\usepackage{stfloats}
\usepackage{tablefootnote}
\usepackage{siunitx}
\usepackage{multirow} % for borders and merged ranges
\usepackage{soul}% for underlines
\usepackage[table]{xcolor} % for cell colors
\usepackage{changepage,threeparttable} % for wide tables

\usepackage[pagebackref,breaklinks,colorlinks]{hyperref}

\usepackage[capitalize]{cleveref}
\crefname{section}{Sec.}{Secs.}
\Crefname{section}{Section}{Sections}
\Crefname{table}{Table}{Tables}
\crefname{table}{Tab.}{Tabs.}

\usepackage{pifont}
\newcommand{\cmark}{\ding{51}}
\newcommand{\xmark}{\ding{55}}

\def\cvprPaperID{*****} % *** Enter the CVPR Paper ID here
\def\confName{CVPR}
\def\confYear{2022}

\begin{document}
	
	%%%%%%%%% TITLE - PLEASE UPDATE
	\title{Long-term Reproducibility for Neural Architecture Search}
	\author{David Towers, Matthew Forshaw, A. Stephen McGough\\
		Newcastle University, UK\\
		{\tt\small \{d.towers2, matthew.forshaw, stephen.mcgough\} }\\
		{\tt\small@newcastle.ac.uk}
		\and
		Amir Atapour-Abarghouei\\
		Durham University, UK\\
		{\tt\small amir.atapour-abarghouei} \\
		{\tt\small @durham.ac.uk}
	}
	\maketitle
	
	%%%%%%%%% ABSTRACT
	\begin{abstract}
	
	It is a sad reflection of modern academia that code is often ignored after publication -- there is no academic `kudos' for bug fixes / maintenance. Code is often unavailable or, if available, contains bugs, is incomplete, or relies on out-of-date / unavailable libraries. This has a significant impact on reproducibility and general scientific progress. Neural Architecture Search (NAS) is no exception to this, with some prior work in reproducibility. However, we argue that these do not consider long-term reproducibility issues. We therefore propose a checklist for long-term NAS reproducibility. We evaluate our checklist against common NAS approaches along with proposing how we can retrospectively make these approaches more long-term reproducible.
	
%		The code associated with many academic works is often left in an unmaintained state after the publication of the associated paper. In this state, works are left bug-ridden, host dead links to other resources, or rely on out-of-date or unavailable libraries. Pressure to move on to new works and a lack of incentive to continue working on old ones, as you cannot publish bug fixes, has impacted reproducibility in many fields. Neural Architecture Search is a prime example of this scenario. This can be harmful to scientific progress. To encourage more active care over code longevity, we propose a checklist that we believe will increase and prolong the reproducibility of works in NAS. We then demonstrate how commonly cited works fair against our checklist and discuss how these works can be adapted to provide future reproducibility.
	\end{abstract}
	
	%%%%%%%%% BODY TEXT
	\section{Introduction}
	\label{sec:intro}
	
	The popularity of Deep Learning (DL) has increased significantly over the last decade. It has enticed us with its ability to solve complex problems, such as image classification, without the need for manual feature identification. However, this has been at the cost of now not knowing the `best' neural architecture to use for a given problem.
	
	Neural Architecture Search (NAS) is a promising field which offers greater potential for finding the `best' architecture for a given data set and problem space -- without needing to train all possible architectures and/or the need of a deep learning expert. Historically, NAS has focused on high accuracy; however, more recently, a move to allow a combination of multiple metrics (e.g.,  accuracy, memory footprint, number of parameters) has emerged \cite{cai_proxylessnas_2019, lu_neural_2021, elsken_efficient_2019}.

%	Prior studies have explored the reproducibility of scientific outputs in several contexts. Hoefler~\emph{et al} conducted a study of the reproducibility of statistical processes in Systems research~\cite{hoefler2015scientific}. Further studies have explored the impact of link rot~\cite{pepe2014astronomers,mayer2015quantitative} and decay in scientific workflows~\cite{zhao2012workflows}. To our knowledge, similar scrutiny has not been applied in the context of Neural Architecture Search.
	
	The potential for NAS was demonstrated by Zoph and Le \cite{zoph_neural_2017} where they outperformed all but one of the human-designed networks they compared against. This potential was limited due to the number of GPU hours it took to find a solution -- which has become the primary driver for future NAS research. New algorithms have reduced computation time -- achieving orders of magnitude improvement \cite{zoph_learning_2018, pham_efficient_2018, liu_darts_2019, cai_proxylessnas_2019, lu_neural_2021, li_block-wisely_2020, chen_drnas_2021, mellor_neural_2021, geada_bonsai-net_2020, geada_spidernet_2022}. However, as the authors move on, their code has been left unmaintained. This is problematic, as it makes it more difficult to use these codes for other data sets and problem spaces, and for comparison with new approaches.
	
	However, there is little incentive to spend time updating old code. While an author could, every few years, spend time updating their previous projects to work with the updated libraries (a task with no academic reward and eventually all-time-consuming), they instead use this time to work on new research, which is more likely to lead to citations. Often driven by the infamous academic mantra of ``Publish or Perish''~\cite{de_rond_publish_2005}. This pressure can take away time academics would otherwise use for activities that do not result in publications, such as teaching students~\cite{rawat_publish_2014} or maintaining code. This neglect can leave codes which are bug-ridden, lacking necessary documentation, or simply out-of-date.
	
    These issues lead to codes that cannot be used without being at least partially re-written. This problem hinders further research based on these codes, removes the ability for others to use the approach for real-world applications and prevents the purpose of the field, which is to replace manually-designed architectures.
	
	Some prior work has looked at the problem of reproducibility~\cite{lindauer_best_2020, pineau_improving_2020} by  proposing a checklist to improve reproducibility of work at the time of publication. This has focused on standards, making sure that the code is complete and that all dependencies are resolvable. However, this only ensures that the work is reproducible at the time of publication. If we seek to make the work reproducible many years after publication, this brings in an additional set of constraints, which need to be considered. These include the long-term availability of dependencies for the code, how changes to libraries effect the running of the code and the availability of a containerised version of the software to encapsulate the code and environment to ensure a longer-term reference used for validation and comparison.
	
	In an ideal world, all authors would adhere to our checklist -- including a containerisation and ease of use. However, this is unlikely to happen for legacy NAS, nor can we expect `usable' systems from new developments. As such, we propose the development of a Docker container approach for legacy NAS approaches and a framework for NAS development separating the NAS approach from core tasks such as dataset input and comparison.
	%---------------------------------------------------------------------------------
	\section{Related Work}
	
	\noindent\textbf{Accessibility of Software Artifacts}
	Many pieces of work are missing software artefacts, hosting dead links, or linking to bug-ridden code. Heumüller~\emph{et al}. show that of 604 reviewed papers that contain mention of software artefacts between 2007 and 2017, only 289 (47.8\%) shared links to these artefacts, and only 163 (27\%) were available. However, they do also show an increase in software artefact availability during this period \cite{heumuller_publish_2020}. This may suggest a more significant push for reproducibility in recent years. 

	This increase is due to several initiatives to encourage artefact sharing, such as ACM badging, including three categories of badges: Artifacts Evaluated, Artifacts Available, and Results Validated, awarded to papers based on a set of criteria. While there is not enough evidence to suggest a cause-and-effect relationship between these initiatives and the increase in software artefact availability, there is enough to suggest these initiatives have influenced this trend \cite{childers_artifact_2018}.

	However, these initiatives do not include prolonged maintenance. Artefact Available badges are awarded for the artefact having persistent storage and an available link. The Artefact Evaluated is awarded for the quality of the artefact at the time of the award. There are no criteria for bug fixing, providing pseudocode, or ensuring usable code if the dependencies are no longer available.
	\newline{}\newline{}\noindent\textbf{Reproducibility in NAS}
	Even when code is made available, there are still problems with reproducibility. We should be able to use virtualisation tools and settable (random) seeds to reproduce the exact experimental conditions. Li and Talwalker found that none of the twelve papers published between 2018-2019 that produced novel NAS works had included their experimental seeds \cite{li_random_2019}, highlighting historical threats to reproducibility in NAS.
	\newline{}\newline{}\noindent\textbf{Reproducibility Checklists}
	Lindauer and Hutter highlight difficulties in reproducing NAS algorithms. They have identified many pitfalls that authors can fall into when publishing their work. They offer a checklist of best practices authors can follow to reduce the chance of succumbing to these problems \cite{lindauer_best_2020}. However, the points raised in their work are for reproducibility at the time of publication but does not encourage practices to prolong reproducibility.
	
	The machine learning checklist\footnote{\href{https://www.cs.mcgill.ca/~jpineau/ReproducibilityChecklist.pdf}{https://www.cs.mcgill.ca/\textasciitilde  jpianeau/ReproducibilityChecklist.pdf}} by Pineau {\em et al.}~\cite{pineau_improving_2020} is adopted by  both ICML and NeurIPS. This checklist has several sections covering all aspects of reproducibility. One of these sections concerns related code and contains points regarding pre-trained models, code availability, and good documentation practices. While this work encourages reproducibility by asking for the code and a list of dependencies, it does not require the code to be bug-free, and if one of those dependencies becomes inaccessible, the code will no longer work.
	
	While we agree with all points in both works, we believe they do not go far enough regarding software artefacts. While they include points about ensuring the code is released, there is little in making sure the code is usable and future-proofed. Following all their checkpoints could still result in code that does not work in a few years. Our checklist is designed to encourage prolonged reproducibility, and hopefully become the basis of a unifying NAS framework which will allow the works to be run into the future.
	%---------------------------------------------------------------------------------
	\section{Reproducing Current Works}
	\label{rcw}
	
	During our investigations into a range of common NAS algorithms\cite{liu_darts_2019, pham_efficient_2018, chen_drnas_2021, xu_pc-darts_2020}, we have found that it is not always a simple and easy process to get them working. Some algorithms still require infeasibly large amounts of computational power to complete such as the original NAS \cite{zoph_neural_2017}. Others need specific versions of libraries or code updating to run. 75\% of the algorithms we have attempted to get working have had some significant issues.
	
	DARTS~\cite{liu_darts_2019} for example, was one of the simplest to reproduce. However, they state\footnote{\href{https://github.com/quark0/darts}{https://github.com/quark0/darts}} a dependency PyTorch 0.3.1. This is technically available, though only for MacOS\footnote{\href{https://pytorch.org/get-started/previous-versions/}{https://pytorch.org/get-started/previous-versions}}. In order to mitigate this, certain sections of DARTS were updated to PyTorch 1.0+ (specific changes on our GitLab page\footnote{\label{mgl}\href{https://gitlab.com/D-Towers/nas-reproducibility}{https://gitlab.com/D-Towers/nas-reproducibility}}). These are the sort of long-term issues which are not as obvious at the time of publication.

	ENAS~\cite{pham_efficient_2018} is a more difficult case. On its GitHub page\footnote{\href{https://github.com/melodyguan/enas}{https://github.com/melodyguan/enas}}, they note the Language Model implementation is now hosted elsewhere. The Image Recognition implementation remains in the repository but lacks a requirements list in the documentation. It can be found, however, in an answered issue. There were also bugs in the code. Community members have fixed these bugs in the issues section. However, these fixes were not pushed to the main branch. Furthermore, ENAS needs a particular version of the \texttt{cudatoolkit} and \texttt{CuDNN} libraries. One of which is archived and needs an NVIDIA account to access, and then needs to be manually installed.
	
    Containerisation tools, such as Docker, can alleviate some of these problems. A properly set-up Dockerfile will recreate the developer environment, so a dependency list would not be required as Docker would handle this automatically. Docker can also be used to save an image of a container, providing a permanent working environment. While this increases the file size considerably, it has several advantages of providing a full set of dependencies, a known working environment and tool, along with removing the issue of dependencies no longer being available.

	%------------------------------------------------------------------------
	
\begin{table*}
    \begin{tabularx}{\textwidth}{@{} l X c c c c @{}}
        \toprule
        Criteria   & Descriptor & \rotatebox{90}{ENAS} & \rotatebox{90}{DARTS} & \rotatebox{90}{PC-DARTS} & \rotatebox{90}{DrNAS} \\ \midrule
        \textbf{Code Stability} &     &      &    &     &      \\               
        Seed Reporting & The code should return the seed used in the logs/output. & \xmark    & \cmark     & \cmark        & \cmark     \\
        Seed Setting & The seed should be able to be set. & \xmark     & \cmark      & \cmark         & \cmark      \\
        Bug-Free   & The code should be free of bugs which prevent sections of code from working. & \xmark     & \cmark      & \cmark         & \xmark\textsuperscript{i}      \\ \midrule
        \textbf{Documentation} &     &      &         &    &  \\               
        Examples & There should be examples of all the high-level commands. & \cmark    & \cmark     & \cmark        & \cmark     \\
        Argument Details & The documentation should include details regarding additional arguments that can be applied to the high-level commands. & \xmark      & \xmark      & \xmark         & \xmark      \\
        Dependency List & Environment specification indicating how to replicate the environment. & \xmark     & \cmark      &    \cmark      & \cmark      \\
        Pipeline Instructions   & Instructions on how to search, train, and test a model from scratch. & \xmark\textsuperscript{ii}      &  \xmark\textsuperscript{iii}     & \xmark\textsuperscript{iii}         & \xmark\textsuperscript{iii}      \\ \midrule
        \textbf{Ease of Running} &     &      &    &     &      \\               
        Dependency Resilience & Dependencies should be installed with the code. & \xmark    & \xmark     & \xmark        & \xmark    \\
        Executable & The code can be executed from an executable file, such as a bash script. & \cmark     & \xmark      & \xmark         & \xmark      \\
        Intuitive Commands   & The commands should be indicative to their function. & \cmark     & \cmark      & \cmark          & \cmark      \\ \midrule
        \textbf{Standardisation} &     &      &     &    &      \\               
        Standard Phases & Search, Train, and Test. & \xmark    & \cmark     & \cmark        & \xmark     \\
        Data Inclusive   & Data is retrieved from a directory named \texttt{data} that accepts common data formats. & \xmark     & \cmark      & \cmark         & \cmark       \\ 
        Standard Output   & The model is saved in an output folder in a form that is easily imported into other scripts. &  \xmark    & \cmark      & \cmark         & \cmark      \\ \midrule
        \textbf{Future Proofing} &     &      &    &     &      \\               
        Accessible Pseudocode & Pseudocode should be made available to allow recreation of the code if all other methods are no longer available. & \xmark    & \xmark     & \xmark        & \xmark     \\
        Containerised Environment & A compressed file of the containerised image (such as tar), which allows the entire environment to be saved, dependencies included. & \xmark & \xmark & \xmark & \xmark \\
        Container Build File & A file containing the required info for a container manager to build the environment and run the code. & \xmark & \xmark & \xmark & \xmark \\
        Source Code   & Source code should be publicly accessible to allow others to inspect and change local versions. & \cmark & \cmark & \cmark & \cmark \\ 
        \bottomrule
    \end{tabularx}
    \caption{Evaluated NAS algorithms using our proposed checklist. \textsuperscript{i}Code is missing a `hp', which needs to be manually entered. \textsuperscript{ii} Does not mention that you need to change the default architecture in the training file. \textsuperscript{iii} Does not explicitly explain how to train a found architecture.}
    \label{table:ourcheck}
\end{table*}
	
%================================

	\section{Proposed Checklist}
	We propose a checklist containing five categories which combines desired features we believe could help increase reproducibility in NAS. Some points in the checklist may become superseded with the inclusion of others. This redundancy should maintain reproducibility if some of the other aspects are not met.
	
	In comparison to the already-existing checklists~\cite{lindauer_best_2020, pineau_improving_2020}, we have focused on aspects which we believe will help prolong reproducibility, with points such as dependency resilience, which ensures the code will remain usable even if a dependency is no loner available.
	
	\Cref{table:ourcheck} shows our checklist applied to four common NAS approaches (ENAS~\cite{pham_efficient_2018}, DARTS~\cite{liu_darts_2019}, PC-DARTS~\cite{xu_pc-darts_2020} and DrNAS~\cite{chen_drnas_2021}). The evaluated approaches have not been future-proofed well, as can be seen from the table. This is evidenced by DARTS requiring a no longer accessible library. We are reproducing several NAS algorithms into a working state in Docker containers. This should allow others to build on our working containers with a working environment. Currently, we have reproduced ENAS into a working Docker Container. These containers will be available, along with any changes we made to the code, on our GitLab\textsuperscript{\ref{mgl}}.
	%-----------------------------------------------------------------------
	\section{Conclusion}
	%Summarise the paper
    In this paper, we have proposed a checklist that is designed to improve and prolong the reproducibility of NAS algorithms. Our checklist covers several areas which can negatively impact how reproducible an author's work is. These areas include code stability, documentation, ease of running, standardisation, and future-proofing.

    In particular, we identify the issue of inaccessible dependencies and suggest containerisation tools as a solution, which allows a researcher's working environment to be saved as an image, including dependencies, which other researchers can import to recreate exactly. 
    
        %Future Work
    We are currently working on recreating the NAS approaches used in this paper into Docker containers, which will be made available on our GitLab\footnote{\href{https://gitlab.com/D-Towers/nas-reproducibility}{https://gitlab.com/D-Towers/nas-reproducibility}}. We have already replicated ENAS into a docker container.
    
    Our repository also includes instructions on how to reproduce environments of each of the NAS approaches used in this paper, using the code from the author's repositories.
    
    We will extend upon this work by developing a framework removing the burden of reproducibility from the NAS developer, thus allowing them to focus on their NAS approach while maintaining a strong level of reproducibility.

	%-----------------------------------------------------------------------
	%%%%%%%%% REFERENCES
	{\small
		\bibliographystyle{ieee_fullname}
		\bibliography{repro}

\begin{thebibliography}{10}\itemsep=-1pt

\bibitem{cai_proxylessnas_2019}
Han Cai, Ligeng Zhu, and Song Han.
\newblock {ProxylessNAS}: {Direct} {Neural} {Architecture} {Search} on {Target}
  {Task} and {Hardware}.
\newblock {\em arXiv:1812.00332 [cs, stat]}, Feb. 2019.

\bibitem{chen_drnas_2021}
Xiangning Chen, Ruochen Wang, Minhao Cheng, Xiaocheng Tang, and Cho-Jui Hsieh.
\newblock {DrNAS}: {Dirichlet} {Neural} {Architecture} {Search}.
\newblock {\em arXiv:2006.10355 [cs, stat]}, Mar. 2021.
\newblock arXiv: 2006.10355 version: 4.

\bibitem{childers_artifact_2018}
Bruce~R. Childers and Panos~K. Chrysanthis.
\newblock Artifact {Evaluation}: {FAD} or {Real} {News}?
\newblock In {\em 2018 {IEEE} 34th {International} {Conference} on {Data}
  {Engineering} ({ICDE})}, pages 1664--1665, Apr. 2018.
\newblock ISSN: 2375-026X.

\bibitem{de_rond_publish_2005}
Mark De~Rond and Alan~N. Miller.
\newblock Publish or {Perish}: {Bane} or {Boon} of {Academic} {Life}?
\newblock {\em Journal of Management Inquiry}, 14(4):321--329, Dec. 2005.

\bibitem{elsken_efficient_2019}
Thomas Elsken, Jan~Hendrik Metzen, and Frank Hutter.
\newblock Efficient {Multi}-objective {Neural} {Architecture} {Search} via
  {Lamarckian} {Evolution}.
\newblock {\em arXiv:1804.09081 [cs, stat]}, Feb. 2019.

\bibitem{geada_spidernet_2022}
Rob Geada and Andrew~Stephen McGough.
\newblock {SpiderNet}: {Hybrid} {Differentiable}-{Evolutionary} {Architecture}
  {Search} via {Train}-{Free} {Metrics}.
\newblock Technical Report arXiv:2204.09320, arXiv, Apr. 2022.
\newblock arXiv:2204.09320 [cs] type: article.

\bibitem{geada_bonsai-net_2020}
Rob Geada, Dennis Prangle, and Andrew~Stephen McGough.
\newblock Bonsai-{Net}: {One}-{Shot} {Neural} {Architecture} {Search} via
  {Differentiable} {Pruners}.
\newblock {\em arXiv:2006.09264}, June 2020.

\bibitem{heumuller_publish_2020}
Robert Heumüller, Sebastian Nielebock, Jacob Krüger, and Frank Ortmeier.
\newblock Publish or perish, but do not forget your software artifacts.
\newblock {\em Empirical Software Engineering}, 25(6):4585--4616, Nov. 2020.

\bibitem{li_block-wisely_2020}
Changlin Li, Jiefeng Peng, Liuchun Yuan, Guangrun Wang, Xiaodan Liang, Liang
  Lin, and Xiaojun Chang.
\newblock Block-{Wisely} {Supervised} {Neural} {Architecture} {Search} {With}
  {Knowledge} {Distillation}.
\newblock In {\em 2020 {IEEE}/{CVF} {Conference} on {Computer} {Vision} and
  {Pattern} {Recognition} ({CVPR})}, pages 1986--1995, Seattle, WA, USA, June
  2020. IEEE.

\bibitem{li_random_2019}
Liam Li and Ameet Talwalkar.
\newblock Random {Search} and {Reproducibility} for {Neural} {Architecture}
  {Search}.
\newblock {\em arXiv:1902.07638 [cs, stat]}, July 2019.
\newblock arXiv: 1902.07638.

\bibitem{lindauer_best_2020}
Marius Lindauer and Frank Hutter.
\newblock Best {Practices} for {Scientific} {Research} on {Neural}
  {Architecture} {Search}.
\newblock Technical Report arXiv:1909.02453, arXiv, Nov. 2020.

\bibitem{liu_darts_2019}
Hanxiao Liu, Karen Simonyan, and Yiming Yang.
\newblock {DARTS}: {Differentiable} {Architecture} {Search}.
\newblock {\em arXiv:1806.09055 [cs, stat]}, Apr. 2019.
\newblock arXiv: 1806.09055.

\bibitem{lu_neural_2021}
Zhichao Lu, Gautam Sreekumar, Erik Goodman, Wolfgang Banzhaf, Kalyanmoy Deb,
  and Vishnu~Naresh Boddeti.
\newblock Neural {Architecture} {Transfer}.
\newblock {\em IEEE Trans. PAMI}, 43(9):2971--2989, Sept. 2021.

\bibitem{mellor_neural_2021}
Joseph Mellor, Jack Turner, Amos Storkey, and Elliot~J. Crowley.
\newblock Neural {Architecture} {Search} without {Training}.
\newblock {\em arXiv:2006.04647 [cs, stat]}, June 2021.

\bibitem{pham_efficient_2018}
Hieu Pham, Melody Guan, Barret Zoph, Quoc Le, and Jeff Dean.
\newblock Efficient {Neural} {Architecture} {Search} via {Parameters}
  {Sharing}.
\newblock In {\em International {Conference} on {Machine} {Learning}}, pages
  4095--4104. PMLR, July 2018.

\bibitem{pineau_improving_2020}
Joelle Pineau, Philippe Vincent-Lamarre, Koustuv Sinha, Vincent Larivière,
  Alina Beygelzimer, Florence d'Alché Buc, Emily~B. Fox, and Hugo Larochelle.
\newblock Improving {Reproducibility} in {Machine} {Learning} {Research} ({A}
  {Report} from the {NeurIPS} 2019 {Reproducibility} {Program}).
\newblock {\em arXiv: Learning}, Mar. 2020.

\bibitem{rawat_publish_2014}
Seema Rawat and Sanjay Meena.
\newblock Publish or perish: {Where} are we heading?
\newblock {\em Journal of Research in Medical Sciences : The Official Journal
  of Isfahan University of Medical Sciences}, 19(2):87--89, Feb. 2014.

\bibitem{xu_pc-darts_2020}
Yuhui Xu, Lingxi Xie, Xiaopeng Zhang, Xin Chen, Guo-Jun Qi, Qi Tian, and
  Hongkai Xiong.
\newblock {PC}-{DARTS}: {Partial} {Channel} {Connections} for
  {Memory}-{Efficient} {Architecture} {Search}.
\newblock {\em arXiv:1907.05737 [cs]}, Apr. 2020.

\bibitem{zoph_neural_2017}
Barret Zoph and Quoc~V. Le.
\newblock Neural {Architecture} {Search} with {Reinforcement} {Learning}.
\newblock {\em arXiv:1611.01578 [cs]}, Feb. 2017.
\newblock arXiv: 1611.01578.

\bibitem{zoph_learning_2018}
Barret Zoph, Vijay Vasudevan, Jonathon Shlens, and Quoc~V. Le.
\newblock Learning {Transferable} {Architectures} for {Scalable} {Image}
  {Recognition}.
\newblock In {\em 2018 {IEEE}/{CVF} {Conference} on {Computer} {Vision} and
  {Pattern} {Recognition}}, pages 8697--8710, Salt Lake City, UT, June 2018.
  IEEE.

\end{thebibliography}
	}
    \clearpage
    
    % Please add the following required packages to your document preamble:
% \usepackage{booktabs}

\end{document}